\documentclass[letterpaper]{article} % DO NOT CHANGE THIS
\usepackage{aaai23}  % DO NOT CHANGE THIS
\usepackage{times}  % DO NOT CHANGE THIS
\usepackage{helvet}  % DO NOT CHANGE THIS
\usepackage{courier}  % DO NOT CHANGE THIS
\usepackage[hyphens]{url}  % DO NOT CHANGE THIS
\usepackage{graphicx} % DO NOT CHANGE THIS
\usepackage{natbib}  % DO NOT CHANGE THIS AND DO NOT ADD ANY OPTIONS TO IT
\usepackage{caption} % DO NOT CHANGE THIS AND DO NOT ADD ANY OPTIONS TO IT
\usepackage{algorithm}
\usepackage{algorithmic}
\usepackage{multirow}
\usepackage{multicol}
\usepackage{diagbox}
\usepackage{amsmath}
\usepackage{amssymb}

\usepackage{newfloat}
\usepackage{listings}
\DeclareCaptionStyle{ruled}{labelfont=normalfont,labelsep=colon,strut=off} % DO NOT CHANGE THIS
\floatstyle{ruled}
\newfloat{listing}{tb}{lst}{}
\floatname{listing}{Listing}
\title{BERT-ERC: Fine-tuning BERT is Enough for Emotion Recognition in Conversation}

\author{
    Xiangyu Qin\textsuperscript{\rm 1,\rm 2}\equalcontrib,
    Zhiyu Wu\textsuperscript{\rm 1}\equalcontrib,
    Jinshi Cui\textsuperscript{\rm 1}\thanks{Corresponding author: cjs@cis.pku.edu.cn},
    Tingting Zhang\textsuperscript{\rm 1},
    Yanran Li\textsuperscript{\rm 2},
    Jian Luan\textsuperscript{\rm 2}\thanks{Corresponding author: luanjian78@hotmail.com},
    Bin Wang\textsuperscript{\rm 2},
    Li Wang\textsuperscript{\rm 3}
    }
\affiliations{
    \textsuperscript{\rm 1}School of Intelligence Science and Technology, Peking University\\
    \textsuperscript{\rm 2}Xiaomi AI Lab\\
    \textsuperscript{\rm 3}School of Psychological and Cognitive Sciences and Beijing Key Laboratory of Behavior and Mental Health, Peking University\\

    2001213087@stu.pku.edu.cn, wuzhiyu@pku.edu.cn, cjs@cis.pku.edu.cn,  zhangtingting3412@gmail.com, yanranli.summer@gmail.com,  luanjian78@hotmail.com,
    wangbin11@xiaomi.com, liwang@pku.edu.cn
}

\begin{document}

\maketitle

\begin{abstract}
Previous works on emotion recognition in conversation (ERC) follow a two-step paradigm, which can be summarized as first producing context-independent features via fine-tuning pretrained language models (PLMs) and then analyzing contextual information and dialogue structure information among the extracted features. However, we discover that this paradigm has several limitations. Accordingly, we propose a novel paradigm, i.e., exploring contextual information and dialogue structure information in the fine-tuning step, and adapting the PLM to the ERC task in terms of input text, classification structure, and training strategy. Furthermore, we develop our model BERT-ERC according to the proposed paradigm, which improves ERC performance in three aspects, namely suggestive text, fine-grained classification module, and two-stage training. Compared to existing methods, BERT-ERC achieves substantial improvement on four datasets, indicating its effectiveness and generalization capability. Besides, we also set up the limited resources scenario and the online prediction scenario to approximate real-world scenarios. Extensive experiments demonstrate that the proposed paradigm significantly outperforms the previous one and can be adapted to various scenes.
\end{abstract}

\section{Introduction}
Emotion Recognition in Conversation (ERC) aims to identify the emotion of each utterance in the dialogue~\cite{Poria2019EmotionRI}. This task has been popularly explored in the NLP research community~\cite{Ghosal2019DialogueGCNAG,Li2021TowardsAO,Gao2021ImprovingER}, which has wide applications in building automatic conversational agents and mining user opinions.

\begin{table}[t]
    \centering
    \begin{tabular}{c|c|c}
    \hline
    \multicolumn{2}{c}{Method} \vline & \multirow{2}*{MELD} \\
    \cline{1-2}
    PLM & classifier & ~ \\
    \hline
    \multirow{5}*{RoBERTa-large} & RGAT & 62.80\\
    ~ & DialogGCN & 63.02\\
    ~ & DAGNN & 63.12\\
    ~ & DialogRNN & 63.61\\
    ~ & DAG-ERC & 63.65\\
    \hline
    RoBERTa-large & MLP & 63.39\\
    \hline
    \end{tabular}
    \caption{Pilot experiment on MELD (\%).}
    \label{tab:toy experiment}
\end{table}

\begin{figure*}
  \centering
  \includegraphics[width=\textwidth]{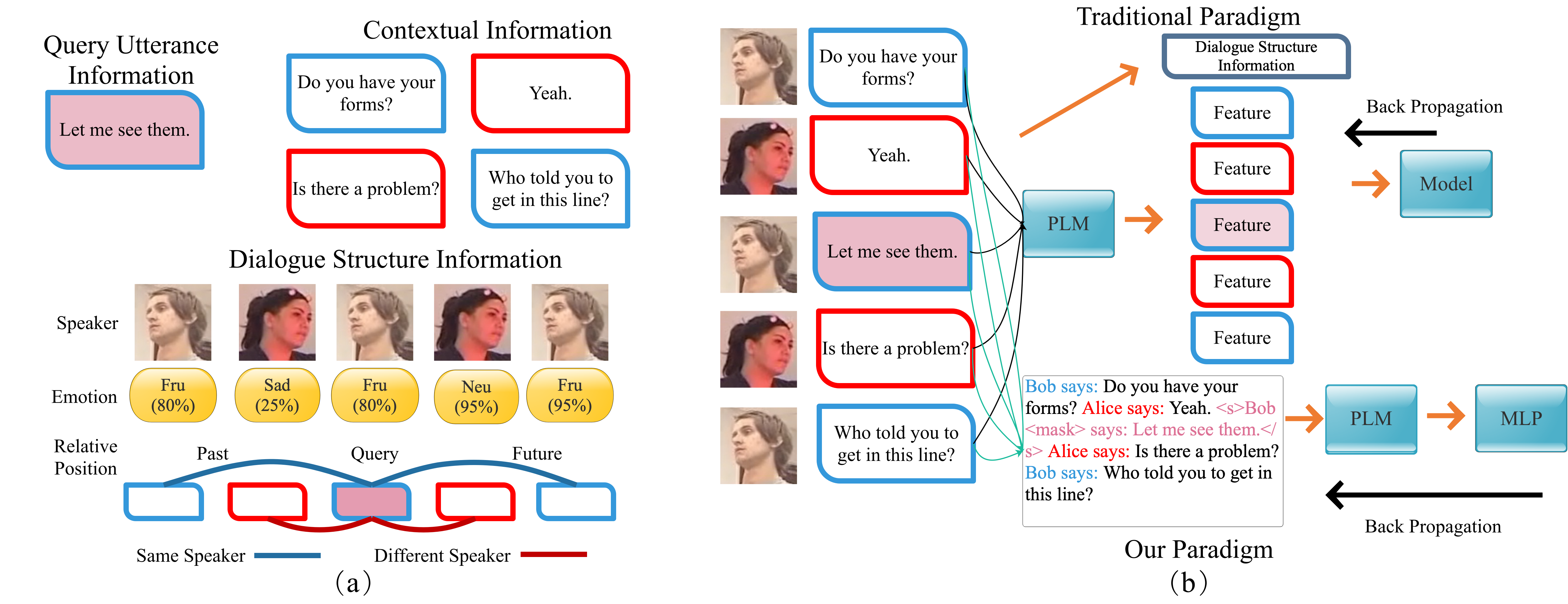}
  \caption{(a) Influencing factors in ERC. (b) Different paradigms for ERC.}
  \label{paradigm}
\end{figure*}

Existing ERC algorithms reveal multiple influencing factors for understanding conversation emotion. As shown in Figure~\ref{paradigm}, we divided these factors into three groups: (1) \textbf{query utterance information} including the text of the query utterance; (2) \textbf{contextual information} including the text of the surrounding utterances (contexts); (3) \textbf{dialogue structure information} consisting of non-textual information of the conversation, such as the speaker information, the emotion states, and the relative position of the utterances. To exploit these three kinds of information, previous works~\cite{ ghosal2019dialoguegcn, majumder2019dialoguernn, ishiwatari2020relation, shen2021directed} commonly follow a two-step paradigm of first extracting context-independent features via fine-tuning pretrained language models (PLMs) and then characterizing contextual information and dialogue structure information among the obtained features by their classifiers (models). For example, DialogRNN~\cite{majumder2019dialoguernn} first extracts utterance features with RoBERTa-large~\cite{liu2019roberta} and then uses three GRUs~\cite{chung2014empirical} to encode contextual information, speaker state, and emotion state, respectively. To verify the contribution of the contextual information and dialogue structure information analyzed in step two, we design a pilot experiment. Specifically, baseline in the experiment uses RoBERTa-large and MLP as the PLM and classifier respectively, suggesting that these two kinds of information remain unexplored. Compared to the methods following the previous paradigm in Table~\ref{tab:toy experiment}, the baseline achieves comparable performance on MELD~\cite{poria2018meld} dataset, indicating that the two kinds of information encoded in the second step only yields trivial improvement (e.g. DialogRNN only outperforms the baseline by 0.22\%). Through analysis, the previous paradigm has two flaws. Firstly, the context-independent features obtained by the PLM are fairly abstract, and thus pose obstacles to analyzing contextual information and dialogue structure information. Secondly, the separation of fine-tuning step and training step leads to extra difficulty in modelling these two kinds of information. Considering these issues, EmoBERTa~\cite{kim2021emoberta} discards the second step and uses entire contexts of the query utterance when fine-tuning. However, it lacks further reflections on the way to adapt the fine-tuning process to the ERC task. Thus, we raise several questions: How to use these three kinds of information when fine-tuning? How to optimize the fine-tuning process according to the characteristics of ERC?

Motivated by these questions, we propose a new paradigm for ERC: integrating query utterance information, contextual information, and dialogue structure information when fine-tuning, and adapting the PLM to the ERC task in terms of input text, classification structure and training strategy. The comparison between the proposed paradigm and the previous one is shown in Figure~\ref{paradigm}. Furthermore, we develop our model BERT-ERC according to the proposed paradigm, which promotes performance in three aspects, namely suggestive text, fine-grained classification module, and two-stage training. (1) Regarding suggestive text, we use the utterances within a certain distance from the query utterance and several indicative tokens, such as speaker name and \textless\textit{mask}\textgreater, to form the input text, thereby indicating speaker information and highlighting the query utterance emotion. (2) The fine-grained classification module considers the temporal structure (past-query-future) of the conversation and generates position-aware features. (3) Concerning the two-stage training, we first train a coarse teacher model via fine-tuning the PLM with the above strategies. Then, we explicitly interpolate the predictions into the input text of the fine student model, allowing it to obtain contextual emotion state. Compared to existing algorithms, both teacher model and student model achieve substantial improvement on four datasets, indicating the effectiveness of these strategies.

In addition to achieving higher accuracy, we note the constraints of ERC in applications, which are ignored by previous works. Thus, we conduct extensive applicability experiments and adapt our paradigm to different scenes. Specifically, we set up the limited resource scenario and the online prediction scenario to approximate real-world scenes. For the former, we design a concise input text structure based on the speaker information to promote the performance of limited-scale PLMs. For the latter, we choose the large-scale PLM and tiny-scale PLM as the coarse teacher and fine student respectively to meet the real-time requirement. 

Overall, our contributions can be summarized as follows: (1) We reveal the limitations of the previous ERC paradigm with a pilot experiment. (2) We propose a new paradigm for ERC: integrating three influencing factors when fine-tuning, and adapting the PLM to the ERC task in terms of input text, classification structure, and training strategy. (3) We develop a new model in three aspects, namely suggestive text, fine-grained classification module, and two-stage training. Moreover, it outperforms existing methods and achieves the accuracy of 71.70\% on IEMOCAP, 67.11\% on MELD, 61.42\% on DailyDialog, 39.84\% on EmoryNLP. (4) We conduct numerous applicability experiments and adapt the proposed paradigm to different scenarios.

\section{Related Work}
\subsection{Emotion Recognition in Conversation}
ERC has received extensive attention in the past decades given its wide applications. Most existing algorithms follow a fixed paradigm that can be generalized as first producing context-independent features and then analyzing contextual information and dialogue structure information. Basically, these methods can be divided into two groups: recurrent-based methods and graph-based methods.

Regarding the recurrent-based methods, HiGRU~\cite{jiao2019higru} uses two GRUs to explore utterance emotion and conversation emotion, respectively. Moreover, DialogRNN~\cite{majumder2019dialoguernn} employs three GRUs to encode context state, speaker state, and emotion state, respectively. COSMIC~\cite{ghosal2020cosmic} is the latest recurrent-based algorithm, which introduces external knowledge into DialogRNN to achieve better performance.

For the graph-based methods, DialogGCN~\cite{ghosal2019dialoguegcn} treats the conversation as a directed graph, where each utterance is connected with the contexts. Differently, DAG-ERC~\cite{shen2021directed} uses a directed acyclic graph to model the dialogue, where each utterance only receives information from the past utterances. Besides, some methods apply Transformer~\cite{vaswani2017attention}, in which self-attention can be viewed as a graph. Specifically, KET~\cite{zhong2019knowledge} combines extra knowledge and transformer encoder to boost performance. DialogXL~\cite{shen2020dialogxl} adapts the transformer to the ERC task via dialog-aware self-attention.

Unlike the above methods, EmoBERTa~\cite{kim2021emoberta} feeds the contexts of the query utterance into the PLM and explores contextual information when fine-tuning.

\subsection{Fine-tuning Methods}
Given the effectiveness of PLMs, researchers typically adapt them to downstream tasks via fine-tuning for better performance. Through our research, existing fine-tuning methods can be divided into three groups: text-based methods, structure-based methods and distillation-based methods.

For text-based methods, Prompt~\cite{kumar2016ask, mccann2018natural,radford2019language, schick2020exploiting} allows the similar structure between the input text of the downstream task and pretraining task. For example, when analyzing the emotion of “Alice: I did well in the exam”, we may attach the prompt “Alice felt \textless\textit{mask}\textgreater”. The \textless\textit{mask}\textgreater \ token enables the PLM to work in a familiar setting and thus improves its performance in the downstream task. Inspired by Prompt, we design the suggestive text, which indicates the dialogue structure via special tokens.

Structure-based methods facilitate fine-tuning mainly in two ways: introducing external knowledge and enabling parameter-efficient transfer learning. For the former, K-BERT~\cite{liu2020k} injects expertise into the PLM by constraining the self-attention module with a knowledge graph. Besides, prefix tuning~\cite{li2021prefix} utilizes a domain word initialized module to emphasize the key content of the downstream task. For the latter, Adapter tuning~\cite{houlsby2019parameter} attaches small neural modules to each layer of the PLM. Moreover, LoRA~\cite{hu2021lora} proposes trainable rank decomposition matrices to reduce trainable parameters. In our work, we design a classification structure based on the characteristics of the ERC task.

Concerning distillation-based methods, they aim to maximally compress PLM size at the cost of limited performance loss. Specifically, TINYBERT~\cite{jiao2019tinybert} proposes a two-stage distillation framework for transformer-based models. Besides, DistilBERT~\cite{sanh2019distilbert} puts forward a lighter BERT~\cite{devlin2018bert} via the knowledge distillation strategy. However, they regard distillation loss as the only knowledge transfer pathway. Unlike the above methods, we shift our focus on promoting the performance of the student model and transfer knowledge via the input text of the student model for the first time.

\section{Methodology}

\begin{figure*}
  \centering
  \includegraphics[width=\textwidth]{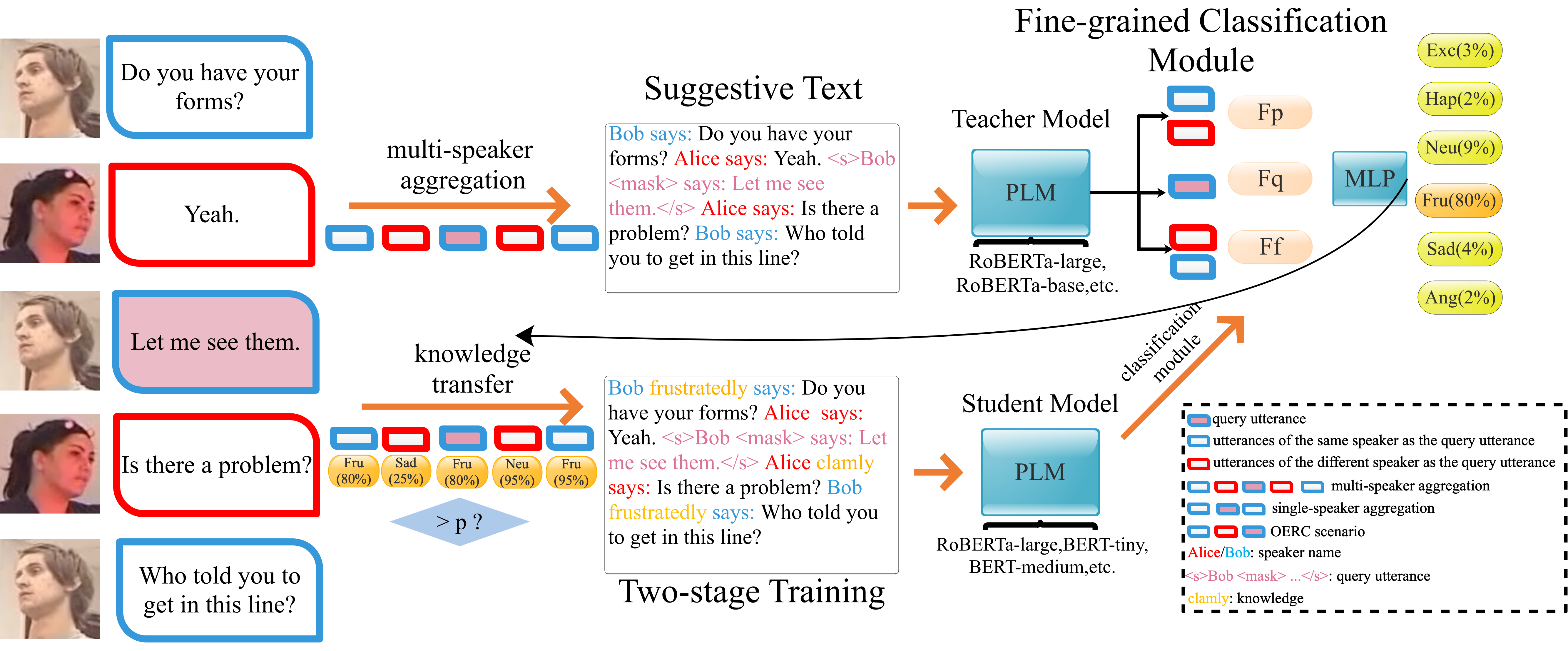}
  \caption{The pipeline of BERT-ERC.}
  \label{method}
\end{figure*}

\subsection{Task Definition}
\begin{figure*}[t]
  \centering
  \includegraphics[width=\textwidth]{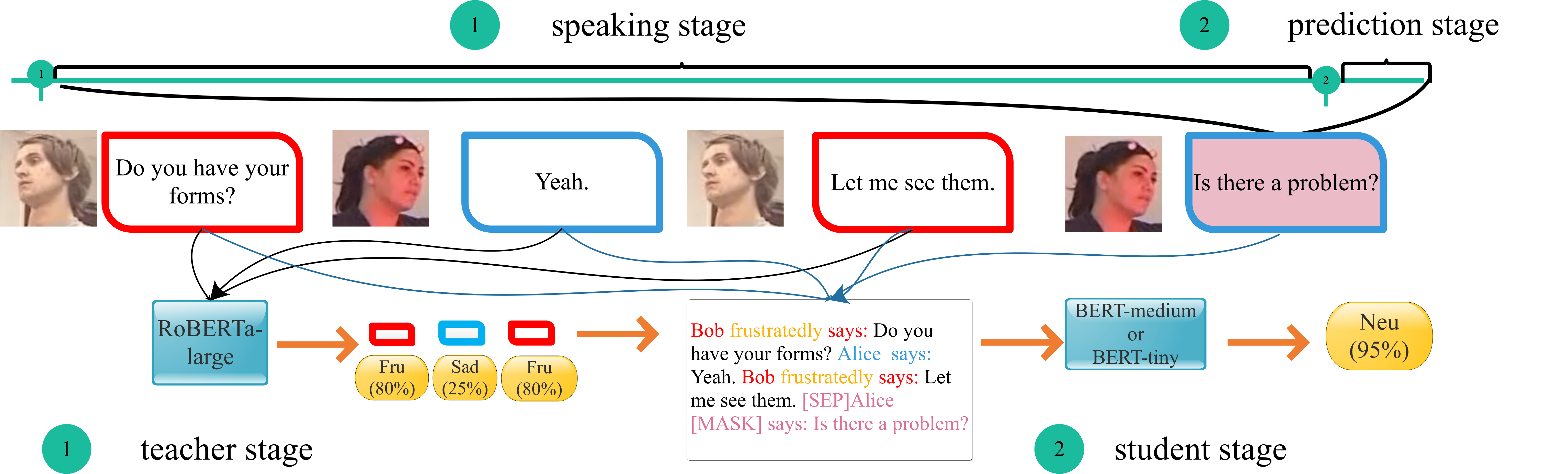}
  \caption{Two-stage training in OERC scenario.}
  \label{oerc}
\end{figure*}

Given a dialogue script along with the speaker information about each constituent utterance, ERC aims to analyze the sentiment of each utterance from a predefined set of emotions. Let $[(u_1, s_1), ..., (u_N, s_N)]$ denote a conversation containing $N$ utterances, where $s_i$ represents the speaker of $u_i$. As illustrated in Figure~\ref{method}, given an utterance $u_i$, the object of ERC is to predict its emotion label $y_i \in Y$ according to the contexts $[u_1, ..., u_N]$ and the corresponding speaker information, where $Y$ denotes the emotion set. In addition to the offline prediction, we also investigate online ERC (OERC) for practical needs. As shown in Figure~\ref{oerc}, given an utterance $u_i$, OERC predicts the emotion label $y_i \in Y$ based on the preceding utterances $[u_1, ..., u_i]$ and speaker information.

In this paper, we define the ERC task as $P(Y|X, M, S)$, where $Y, X, M, S$ denote predictions, input text generation approach, PLM, and training strategy respectively. Furthermore, we design a new paradigm for ERC, which can be summarized as integrating three influencing factors (query utterance information, contextual information, and dialogue structure information) during fine-tuning, and adapting the PLM to the ERC task in terms of input text, classification structure and training strategy. In other words, we select the most appropriate ($X$, $M$, $S$) in different scenarios. According to the proposed paradigm, we develop our model BERT-ERC in three aspects: suggestive text, fine-grained classification module, and two-stage training. Details of these strategies will be presented as follows. 

\subsection{Suggestive Text}
Let $[(u_1, s_1), ..., (u_N, s_N)]$ denote the conversation, and $x_i$ represents the input text of the query utterance $u_i$. Traditional algorithms only feed the query utterance into the PLM, i.e., $x_i = u_i$, which proved to be a suboptimal strategy. Thus, we use utterances within a certain distance from the query utterance to form the input text. Nonetheless, directly splicing different utterances probably yields negligible improvement, as the PLM comprehends limited knowledge of dialogue structure in pretraining. Accordingly, we explicitly introduce dialogue structure information and contextual information into the input text:
\begin{align}
    \begin{split}
        x_i = & [X_p(u_a, s_a), ..., X_p(u_{i-1}, s_{i-1}), \\ & \ X_q(u_i, s_i), \\ & \ X_f(u_{i+1}, s_{i+1}), ..., X_f(u_b, s_b)]
    \end{split}
\end{align}
where $a$, $b$ denote the range of the contexts, $X_p$, $X_q$, and $X_f$ denote the corresponding strategy for past utterances, query utterance, and future utterances. We provide an example in Figure~\ref{method}, and details will be presented as follows.

Regarding $X_q$, we exploit three kinds of special tokens. Firstly, we place \textit{speaker says:} ahead of the query utterance to provide speaker information. Secondly, \textless\textit{s}\textgreater \ and \textless\textit{/s}\textgreater \ are employed to enclose the query utterance for emphasis. Thirdly, we apply the \textless\textit{mask}\textgreater \ token to focus the model on the emotion state of the query utterance. In other words, the suggestive query utterance can be expressed by:
\begin{align}
    X_q(u_i, s_i) \ = \ \textless\textit{s}\textgreater s_i \  \textless\textit{mask}\textgreater \ \textit{says: } u_i\textless\textit{/s}\textgreater
\end{align}
For $X_p$ and $X_f$, \textit{speaker says:} serves as the only indication given the supporting role of the contexts:
\begin{align}
    X_p(u_j, s_j) \ &= \ s_j  \textit{ says: } u_j \\
    X_f(u_j, s_j) \ &= \ s_j \textit{ says: } u_j
\end{align}
We denote the above method as multi-speaker aggregation, as it involves all contexts within a certain distance from the query utterance. However, limited-scale PLMs still perform poorly in modelling contextual information and dialogue structure information even with these special tokens. Considering that utterances of the same speaker as the query one can better reflect the emotion state of the speaker, we propose single-speaker aggregation, thereby reducing the task to exploring the mood swings of a specific speaker. Formally, let $X_p^s$, $X_q^s$, and $X_f^s$ denote the corresponding strategy for the three kinds of utterances in single-speaker aggregation. Thus, the aggregated input text can be expressed by:
\begin{align}
    X_q^s(u_i, s_i) \ &= \ \textless\textit{s}\textgreater s_i \  \textless\textit{mask}\textgreater \ \textit{says: } u_i\textless\textit{/s}\textgreater \\
X_p^s(u_j, s_j) \ &= \ s_j \textit{ says: } u_j \ \ \textit{ if }s_j == s_i \textit{ else None}  \\
X_f^s(u_j, s_j) \ &= \ s_j \textit{ says: } u_j \ \ \textit{ if }s_j == s_i \textit{ else None} 
\end{align}

\subsection{Fine-grained Classification Module}
Given the input text $x_i$ of the query utterance $u_i$, PLM generates the features of each constituent token. Traditional fine-tuning methods generally utilize the \textit{class} token for classification, as the input of most tasks is a piece of text without any special structures. However, $x_i$ has principal-subordinate structure (query utterance in leading position, contexts in supporting status) and temporal structure (past-query-future), indicating that the previous classifier is suboptimal for our model. Accordingly, we propose a fine-grained classification module according to the traits of ERC, whose details will be presented as follows.

Let $[f_1, ..., f_l]$ denote the features of $x_i$, where $[f_a, ..., f_b]$ correspond to the query tokens. We first divide the tokens into past tokens, query tokens, and future tokens based on the position. Then, past features $F_p$, query features $F_q$, and future features $F_f$ are generated via mean operation:
\begin{align}
    F_p \ &= \ \textit{mean}([f_1, ..., f_{a-1}]) \\
    F_q \ &= \ \textit{mean}([f_a, ..., f_b]) \\
    F_f \ &= \ \textit{mean}([f_{b+1}, ..., f_l])
\end{align}
Afterwards, we get the concatenated feature $F_{cls}=[F_p, F_q, F_f]$, which contains both principal-subordinate structure information and temporal structure information.

Similar to the processing of the \textit{class} token, a fully connected layer followed by the \textit{Tanh} activation function is utilized for projection. Finally, we use the Dropout~\cite{srivastava2014dropout} layer to prevent overfitting and the MLP for classification. In other words, prediction $\hat{y_i}$ can be computed by:
\begin{align}
    \hat{y_i} = \textit{MLP}(\textit{Dropout}(\textit{Tanh}(\textit{FC}(F_{cls}))))
\end{align}

\subsection{Two-stage Training}
Given the significance of contextual emotion state, existing algorithms implicitly exploit it via modelling dialogue structure information. Differently, the proposed paradigm makes it possible to explicitly introduce contextual emotion state into the input text. Inspired by knowledge distillation~\cite{hinton2015distilling}, we utilize the coarse teacher - fine student framework. Specifically, we first train a teacher model via fine-tuning the PLM with aforementioned strategies and then interpolate the predictions with high confidence into the input text of the student model. It is worth noting that the two-stage training strategy imposes no constraints on the two models. Accordingly, we test various combinations of PLMs to meet the requirements of different scenarios. The pipeline of the two-stage training is shown in Figure~\ref{method}, and we will illustrate it in terms of knowledge, framework, and combination of different PLMs.

\subsubsection{Knowledge}
According to the emotion set $Y$ of the teacher model, we divide the knowledge into task driven knowledge and common-sense-based knowledge. For the former, $Y$ includes all classes involved in the task. In other words, the teacher model first completes the task and then transfers the predictions with high confidence as the knowledge to the student model. For the latter, we draw inspiration from the way humans perceive emotion state. Specifically, given the predominance of \textit{neutral} emotion in daily conversations, people first simplify ERC as a binary (neutral, emotional) or ternary (positive, neutral, negative) problem, and then conduct further classification. Similarly, the teacher model completes a simplified ERC task and then imparts the predictions to the student model to conduct fine-grained classification.

\subsubsection{Framework}
In two-stage training, we follow the coarse teacher - fine student framework. Firstly, the teacher model generates a pseudo label for each utterance and filters out the low confidence predictions. For a given dialogue $[(u_1, s_1), ..., (u_N, s_N)]$, let $\hat{y_i}$ denotes the prediction of $u_i$ and $p_i$ represents the prediction confidence of $\hat{y_i}$. Thus, the utterance $i$ embedded with knowledge can be expressed by $f(u_i, s_i, \hat{y_i}, p_i)$, and $f$ denotes the screening strategy:
\begin{align}
f(u_i, s_i, \hat{y_i}, p_i)=
\begin{cases}
(u_i, s_i, \hat{y_i}) & \textit{ if } p_i \geq p \\
(u_i, s_i) & \textit{ otherwise}
\end{cases}
\end{align}
where $p$ is a hyperparameter in our model. Secondly, we introduce the knowledge into the input text of the student model and change the suggestive text strategy of the student:
\begin{align}
    X_q(u_i, s_i, \hat{y_i}) \ &= \ \textless\textit{s}\textgreater s_i \  \textless\textit{mask}\textgreater \ \textit{says: } u_i\textless\textit{/s}\textgreater \\
    X_p(u_i, s_i, \hat{y_i}) \ &= \ s_i \ \textless\textit{emo}\textgreater \textit{says: } u_i \\
    X_f(u_i, s_i, \hat{y_i}) \ &= \ s_i \ \textless\textit{emo}\textgreater \textit{says: } u_i
\end{align}
where \textless\textit{emo}\textgreater \ corresponds to the emotion label of $\hat{y_i}$. For example, we set \textless\textit{emo}\textgreater \ to \textit{angrily} if and only if $\hat{y_i}$ denotes \textit{anger}. In such a manner, the explicitly indicated contextual emotion states improve the performance of the student model. In test phase, we first make predictions with the teacher model and then generate the knowledge, which will be imparted to the student for refined predictions.

\subsubsection{Combination of Different PLMs}
As mentioned above, we choose the combination of PLMs according to the experimental scenario. To achieve optimal accuracy, we use RoBERTa-large as the PLM of both teacher and student. However, large-scale PLMs cannot meet the real-time requirement of OERC, which is also a common issue in prevailing algorithms. To solve this problem, we divide OERC into the speaking stage and the prediction stage. The model is unoccupied in the former stage (3-5 seconds), as it needs the query utterance for emotion recognition. For the latter, the model is required to assess the emotion in 50-100 milliseconds. Considering the long duration of the first stage, the teacher-student framework perfectly fits OERC scenario. Specifically, in the speaking stage, a large-scale PLM is used to generate past emotion states. Then, in the prediction stage, we use a tiny-scale PLM as the student to conduct online prediction based on the text embedded with knowledge.

\section{Experiments}

\begin{table*}[t]
    \centering
    \begin{tabular}{c|c c c c}
    \hline
    Method & IEMOCAP & MELD & DailyDialog & EmoryNLP \\
    \hline
    DialogRNN + RoBERTa~\cite{majumder2019dialoguernn} & 64.76 & 63.61 & 57.32 & 37.44 \\
    DialogGCN + RoBERTa~\cite{ghosal2019dialoguegcn} & 64.91 & 63.02 & 57.52 & 38.10  \\
    RGAT + RoBERTa~\cite{ishiwatari2020relation} & 66.36 & 62.80 & 58.08 & 37.78 \\
    KET~\cite{zhong2019knowledge} & 59.56 & 58.18 & 53.37 & 33.95 \\
    DialogXL~\cite{shen2020dialogxl} & 65.94 & 62.41 & 54.93 & 34.73\\
    DAGNN~\cite{shen2021directed} & 64.61 & 63.12 & 58.36 & 37.98 \\
    COSMIC~\cite{ghosal2020cosmic} & 65.28 & 65.21 & 58.48 & 38.11 \\
    DAG-ERC~\cite{shen2021directed} & 68.03 & 63.65 & 59.33 & 39.02 \\
    \hline
    EmoBERTa~\cite{kim2021emoberta} & 68.57 & 65.61 & - & - \\
    \hline
    CoMPM~\cite{lee2021compm} & 69.46 & 66.52 & 60.34 & 38.93 \\
    T-GCN~\cite{lee2021graph} & - & 65.36 & \textbf{61.91} & 39.24 \\
    \hline
    BERT-ERC (teacher) & 69.43 & 66.15 & 60.71 & 39.73\\
    BERT-ERC (student) & 70.84 & 66.65 & 61.42 & 39.84\\
    BERT-ERC (best) & \textbf{71.70} & \textbf{67.11} & 61.42 & \textbf{39.84}\\
    \hline
    \end{tabular}
    \caption{Comparison with the state-of-the-art methods on four datasets (\%).}
    \label{tab:comparison with sota}
\end{table*}

\subsection{Datasets}
Our experiments involve four datasets, whose information is as follows. \textbf{IEMOCAP}~\cite{busso2008iemocap} is a multi-modal dataset, where each utterance is labelled with one of six emotions, namely \textit{neutral, happiness, sadness, anger, frustrated, and excited}. Following previous works, dialogues of the first four sessions are used as the training set and the rest are used as the test set. \textbf{MELD}~\cite{poria2018meld} is a multi-modal dataset extracted from the TV show \textit{Friends}. It contains seven emotion labels: \textit{anger, disgust, fear, happiness, sadness, surprise, and neutral}. \textbf{DailyDialog}~\cite{li2017dailydialog} collects conversations of English learners. It includes the same seven emotion labels as MELD. \textbf{EmoryNLP}~\cite{zahiri2018emotion} is also built on the TV show \textit{Friends}, but differs from MELD in scenes and labels. It contains seven types of labels: \textit{neutral, sad, mad, scared, powerful, peaceful, and joyful}. In our experiments, we only utilize textual modality. Regarding evaluation metrics, we follow previous works and choose micro-averaged F1 excluding \textit{neutral} for DailyDialog and weighted-average F1 for the rest datasets. 

\subsection{Implementation Details}
We conduct experiments in three application scenarios. For offline prediction (Section 4.3), we fix all parameter settings and assess our model on four datasets. Moreover, in Section 4.4 and 4.5, we conduct experiments on IEMOCAP in limited resources scenario and OERC scenario to approximate real-world scenes respectively. %Besides, we discuss the methods for adapting each component of BERT-ERC to different datasets in the Appendix A.1 and A.2. 
Details are as follows.

Regarding \textbf{offline prediction}, the proposed model predicts the emotion of each utterance with sufficient time, space and computational resources. To achieve the optimal performance, we use RoBERTa-large~\cite{liu2019roberta} with the first 8 encoder layers frozen as the PLM. Moreover, we utilize the multi-speaker aggregation and set the knowledge confidence $p$ to $0.7$. For the \textbf{limited resources scenario}, such as mobile devices, due to the limitations of space and computational resources, we have to use limited-scale models. Besides, we expect more frozen parameters when fine-tuning for parameter reuse. Thus, we choose RoBERTa-base~\cite{liu2019roberta} with the first 6 or 10 encoder layers frozen as the PLM. Concerning \textbf{OERC}, to exploit the time in the speaking stage, we use RoBERTa-large with the first 8 layers frozen as the PLM of the teacher. Besides, we employ BERT-tiny~\cite{turc2019well} and BERT-medium~\cite{turc2019well} as the PLM of the student. Moreover, we set $p$ to $0.5$.

Scenario-independent settings are listed as follows. We truncate the input text to meet the requirement of PLMs. We use the Adam optimizer~\cite{kingma2014adam} with a learning rate of 9e-6 in experiments. Besides, we utilize a 1-layer MLP as the classifier unless otherwise specified. For all datasets, we train 10 epochs with the batch size of 8. Focal Loss~\cite{lin2017focal} is applied to alleviate the class imbalance problem. We implement all experiments on 4 NVIDIA Tesla V100 GPUs with the Pytorch framework.

\subsection{Offline Prediction Scenario}
\subsubsection{Comparison with the State-of-the-Art Methods}
We compare BERT-ERC with several state-of-the-art methods on four datasets in Table~\ref{tab:comparison with sota}. The first 8 lines present the performance of several traditional algorithms, which model contextual information and dialogue structure information based on the context-independent features. Through comparison, algorithms using contexts when fine-tuning (lines 9-14) outperform traditional methods, suggesting the significance of exploring contextual information and dialogue structure information in the extraction stage. Moreover, compared to EmoBERTa~\cite{kim2021emoberta}, our model achieves substantial improvement as we adapt the fine-tuning process to the ERC task. Concurrently, CoMPM~\cite{lee2021compm} and T-GCN~\cite{lee2021graph} insert contextual information into PLMs by extra models. Nonetheless, BERT-ERC still leads in most cases, which mainly benefits from the suggestive text and the fine classifier. %Furthermore, we choose appropriate knowledge, which will be discussed in Appendix A.1, for each dataset and generate the fine student model in line 11. 
Furthermore, we use the two-stage training strategy to generate the fine student model in line 13, which outperforms the coarse teacher in all datasets, indicating that proper knowledge can further boost the performance of our paradigm. Besides, we use a 2-layer MLP as the classifier on IEMOCAP and freeze more PLM layers on MELD
%, which will be discussed in Section 4.3.2 and Appendix A.2, 
to achieve the optimal performance on these two datasets (line~14). Overall, our model achieves leading performance on four datasets, demonstrating the advantages of the proposed paradigm.

\begin{table*}[t]
    \centering
    \begin{tabular}{c c c c c|c}
    \hline
    \textless\textit{mask}\textgreater & contexts & FCM & 2-layer MLP & two-stage training & IEMOCAP \\
    \hline
     & & & & & 56.96 \\
    \checkmark & & & & & 57.72 \\
    \checkmark & \checkmark & & & & 68.55 \\
    \checkmark & \checkmark & \checkmark & & & 69.43 \\
    \checkmark & \checkmark & \checkmark & \checkmark & & 71.15 \\
    \checkmark & \checkmark & \checkmark & \checkmark & \checkmark & 71.70 \\
    \hline 
    \end{tabular}
    \caption{Ablation study on IEMOCAP (\%). (FCM: Fine-grained classification module)}
    \label{tab:ablation study}
\end{table*}

\subsubsection{Ablation Study}
We design an ablation study on IEMOCAP to diagnose the proposed modules, whose results are shown in Table~\ref{tab:ablation study}. (1) The baseline model uses the query utterance as the input and employs the \textit{class} token for classification. (2) The \textless\textit{mask}\textgreater \ token emphasizes the emotion state of the query utterance and improves the performance by 0.76\%. 
%\textit{More experiments on the} \textless\textit{mask}\textgreater \ \textit{token are shown in Appendix A.3}. 
(3) We get the improvement of 10.83\% via the contexts of the query utterance, demonstrating that exploring contextual information and dialogue structure information in the fine-tuning step is the most critical strategy in ERC. (4) FCM (line 4) further gains the progress of 0.88\%, which can be credited to the introduced dialogue structure. (5) Inspired by ViT~\cite{dosovitskiy2020image}, we believe that 2-layer classification MLP outperforms 1-layer MLP in high quality datasets. Compared to the latter, the former obtains the improvement of 1.72\%. (6) Compared to the teacher, the fine student model achieves the improvement of 0.55\%, suggesting the effectiveness of the two-stage training.

\begin{table}[t]
    \centering
    \begin{tabular}{c|c}
    \hline
    Method & IEMOCAP \\
    \hline
    DialogRNN + RoBERTa-large & 64.76 \\
    DialogRNN + RoBERTa-base & 62.75 \\
    \hline
    DialogGCN + RoBERTa-large & 64.91 \\
    DialogGCN + RoBERTa-base & 64.18 \\
    \hline
    RGAT + RoBERTa-large & 66.36 \\
    RGAT + RoBERTa-base & 65.22 \\
    \hline
    BERT-ERC + RoBERTa-large + MSA & 69.43 \\
    BERT-ERC + RoBERTa-base (fr6) + MSA & 66.87 \\
    BERT-ERC + RoBERTa-base (fr6) + SSA & 68.98 \\
    \hline
    BERT-ERC + RoBERTa-base (fr10) + MSA & 63.22 \\ 
    BERT-ERC + RoBERTa-base (fr10) + SSA & 66.16 \\
    \hline
    \end{tabular}
    \caption{Performance comparison on IEMOCAP in the limited resources scenario (\%). (MSA: multi-speaker aggregation; SSA: single-speaker aggregation; RoBERTa-base (fr$n$): RoBERTa-base with the first $n$ layers frozen;)}
    \label{tab:limited resources}
\end{table}

\subsection{Limited Resources Scenario}
To approximate the scene of conducting ERC on mobile devices, we set up the limited resources scenario and conduct experiments in Table~\ref{tab:limited resources}. According to the first 6 lines, limited-scale PLMs hazard existing algorithms, possibly due to the reduced modelling capability. Besides, our model suffers from a more severe performance drop, as it is built entirely on the PLM. However, BERT-ERC with a small PLM still outperforms most previous works, suggesting that integrating query utterance information, contextual information and dialogue structure information when fine-tuning is the most critical strategy in ERC. Moreover, single-speaker aggregation (line 9) obtains the progress of 2.11\% compared to multi-speaker aggregation (line 8), indicating that focusing only on the mood swings of one speaker improves ERC performance in the resource-limited condition. %\textit{Comparison between these two strategies on all datasets is shown in Appendix E}. 
Besides, we freeze the first 10 layers of RoBERTa-base for more parameter reuse and achieve the accuracy of 66.16\% with single-speaker aggregation, which is comparable to the performance of existing methods. Insofar as we know, we are the first to consider PLM size and parameter reuse in ERC, and results show that our paradigm can be adapted to various scenarios by altering the text generation strategy.

\begin{table}[t]
    \centering
    \begin{tabular}{c|c c}
    \hline
    \diagbox{Method}{Acc (\%)}{PLM} & BERT-tiny & BERT-medium \\
    \hline
    MSA + C & 42.58 & 61.64 \\
    MSA + K + C & 48.22 & \textbf{69.62} \\
    SSA + K + C & 56.27 & 69.02 \\
    SSA + K & \textbf{63.27} & 68.37 \\
    \hline
    \end{tabular}
    \caption{Performance comparison on IEMOCAP in OERC scenario (\%). (MSA: multi-speaker aggregation; SSA: single-speaker aggregation; K: knowledge; C: contexts)}
    \label{tab:OERC}
\end{table}

\subsection{OERC}
In OERC scenario, we choose RoBERTa-large as the coarse teacher and use BERT-tiny and BERT-medium as the fine student. Turc et al.~\cite{turc2019well} point out that these two PLMs are $3 \times$ and $65 \times$ faster than RoBERTa-large in inference respectively. The experimental results are shown in Table~\ref{tab:OERC}. Baseline model does not use the knowledge (line~1) and fails to meet practical needs. Compared to the baseline, two-stage training introduces contextual emotion state as the knowledge (line 2) and promotes the performance by 5.64\% and 7.98\%. It is worth noting that BERT-medium with teacher achieves $3 \times$ faster inference while still outperforming the state-of-the-art methods in offline prediction scenario. To further boost the performance of BERT-tiny, we utilize single-speaker aggregation (line 3) and achieve the improvement of 8.05\%. Furthermore, we simplify the input text by replacing the preceding utterances with the revealed emotion states and achieves an accuracy of 63.27\% with BERT-tiny. Reviewing the strategies in line 2 to 4, we discover that simpler input text advances limited-scale PLMs. However, these two strategies fail to benefit BERT-medium, as it has sufficient parameters to model dialogue structure and past utterances. Overall, the proposed two-stage training strategy yields great inference speed gains with limited accuracy loss and achieves great performance on different PLMs with the corresponding text generation approach.

\section{Conclusion}
Given the flaws of the previous ERC paradigm, we put forward a new one in this paper and further develop the BERT-ERC model according to the proposed paradigm. Our model utilizes three strategies (suggestive text, fine-grained classification module, and two-stage training) to introduce dialogue structure information and contextual information into the PLM. Through extensive experiments, BERT-ERC achieves state-of-the-art performance on four datasets. Besides, we set up the limited resources scenario and OERC scenario to approximate real-world scenes. Overall, comprehensive experiments demonstrate the generalization ability and effectiveness of the proposed paradigm.

\section{Acknowledgments}
We would like to thank the intelligent voice assistant \textit{XiaoAi} and the \textit{Xiaomi AI Lab} for their guidance and inspiration to our emotion conversation work.

\bibliography{aaai23}

\end{document}

\appendix

\section{Component Analysis}
Through extensive experiments, BERT-ERC achieves the best available results on four datasets. In this part, we explore ways to adapt our model to different datasets in terms of knowledge, modelling capability and \textless\textit{mask}\textgreater \ token.

\subsection{Knowledge}

As mentioned in Section 3.4.1, we propose task driven knowledge and common-sense-based knowledge. For class-balanced datasets with consistent distribution between training set and test set, such as IEMOCAP, task driven knowledge outperforms common-sense-based knowledge. Moreover, in EmoryNLP, where the distribution of the training set and test set is inconsistent, we first complete a ternary ERC task to prevent overfitting. Besides, in daily conversations, such as DailyDialog and MELD, where neutral emotion predominates, identifying it first yields optimal performance. Results in Section 4.3.1 indicate that the student with appropriate knowledge achieves great improvement.

\subsection{Modelling Capability}

Traditional ERC algorithms balance overfitting and underfitting by altering the complexity of the model. Differently, in our paradigm, we adjust the modelling capability via changing the number of frozen layers and MLP layers. As illustrated in the ablation study, 2-layer classification MLP outperforms 1-layer MLP in IEMOCAP, as the stronger representation capability contributes to the performance in high quality dataset. In addition, reducing parameters facilitates the generalization ability in low quality datasets. Specifically, in MELD, we freeze the first 16 layers of the RoBERTa-large and obtain the progress of 0.96\% (67.11\%) compared to the RoBERTa-large with the first 8 layers frozen (66.15\%).

\subsection{\textless\textit{mask}\textgreater \ Token}
We have demonstrated the effectiveness of the \textless\textit{mask}\textgreater \ token in the ablation study. Practically, its effect becomes more pronounced as the task becomes more difficult. In Table~\ref{tab:<mask> on EmoryNLP}, we conduct an ablation study of the \textless\textit{mask}\textgreater \ token on EmoryNLP, which serves as the most challenging dataset. As results illustrate, the model confronts instability and fitting problems without the \textless\textit{mask}\textgreater \ token. Moreover, it also plays an important role in OERC. Specifically, the student model without \textless\textit{mask}\textgreater \ token suffers a drop of 1.02\%. Through analysis, the \textless\textit{mask}\textgreater \ token benefits the model in two aspects. Firstly, in two-stage training, it allows the similar sentence structure between the query utterance and the preceding utterances in the student model, thereby emphasizing the past emotion state indicated by the pseudo labels. Secondly, it directly focuses the model on the emotion state of the query utterance and facilitates stable training.

\section{Stability Analysis}
Regarding all the aforementioned experiments, we set the random seed to $0$ so as to facilitate reproduction. In this section, we investigate the stability of the proposed model on IEMOCAP and MELD. Specifically, we choose 5 random seeds (0 - 4) and take the average and standard deviation as the measurement. Moreover, the models (BERT-ERC (teacher), BERT-ERC (student), BERT-ERC (best)) involved in this experiment remain the same as those in Section 4.3.1. As the results in Table~\ref{tab:stability analysis} illustrate, our model consistently outperforms existing ERC algorithms in multiple turns, indicating its effectiveness and stability.

\begin{table}[t]
    \centering
    \begin{tabular}{c|c|c|c|c}
        \hline
        Random Seed & 0 & 1 & 2 & 3 \\
        \hline
        with \textless\textit{mask}\textgreater \ & \textbf{39.73} & 38.46 & \textbf{40.03} & \textbf{39.11} \\
        \hline
        without \textless\textit{mask}\textgreater \ & 10.94 & \textbf{38.47} & 36.97 & 10.94 \\
        \hline
    \end{tabular}
    \caption{Ablation Study (\%) of the \textless\textit{mask}\textgreater \ token on EmoryNLP.}
    \label{tab:<mask> on EmoryNLP}
\end{table}

\begin{table}[t]
    \centering
    \begin{tabular}{c|c|c|c|c}
    \hline
    \multirow{2}*{Methods} & \multicolumn{2}{c}{IEMOCAP} \vline & \multicolumn{2}{c}{MELD} \\
    \cline{2-5}
    ~ & mean & std & mean & std \\
    \hline
    BERT-ERC (teacher) & 70.41 & 0.63 & 66.01 & 0.21 \\
    BERT-ERC (student) & 71.12 &  0.56 & 66.26 & 0.33\\
    BERT-ERC (best) & 71.53 & 0.21 & 66.89 & 0.27 \\
    \hline
    \end{tabular}
    \caption{Stability analysis of BERT-ERC (\%).}
    \label{tab:stability analysis}
\end{table}

\section{Discussion}
\subsection{Time and Computational Resources}
The proposed paradigm utilizes the contexts of the query utterance in the fine-tuning step, thus leading to intensive computation. To solve this problem, we put forward several methods. (1) We demonstrate the effectiveness of single-speaker aggregation in the limited resource scenario in Section 4.4, which enables concise input text. (2) We utilize a large-scale PLM and a tiny-scale PLM as the teacher and student, respectively, to meet the real-time requirement of OERC. However, these methods are not enough to eliminate the computational burden introduced by the contexts. Inspired by the success of the \textit{SSA + K} mode on BERT-tiny in Section 4.6, the future work lies in the following aspects. Firstly, we will explore methods in which the tiny-scale PLM transfers knowledge to itself. Secondly, we will further improve the strategy of replacing the contexts with the pseudo-labels and reduce the performance loss caused by this method. 

\begin{figure*}[t]
  \centering
  \includegraphics[width=\textwidth]{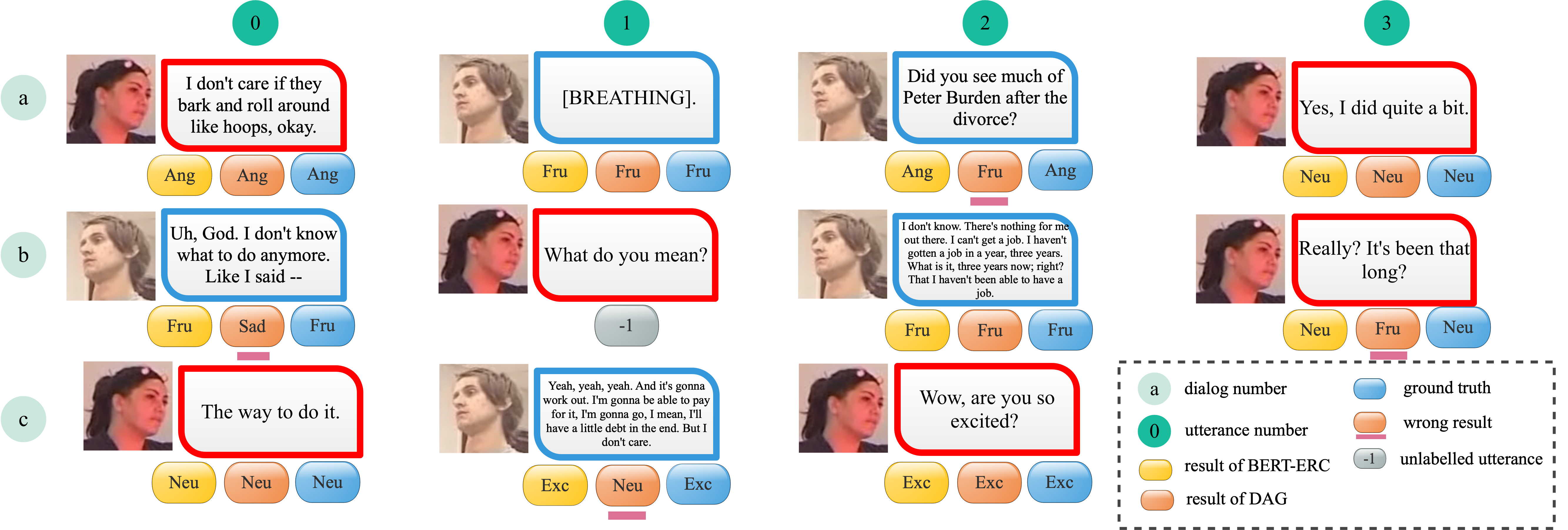}
  \caption{Visualization of several utterances in the IEMOCAP dataset.}
  \label{example}
\end{figure*}

\begin{table*}[t]
    \centering
    \begin{tabular}{c|c|c|c|c|c|c|c|c}
    \hline
    \multirow{2}*{Methods} & \multicolumn{2}{c}{IEMOCAP} \vline & \multicolumn{2}{c}{MELD} \vline & \multicolumn{2}{c}{DailyDialog} \vline &
    \multicolumn{2}{c}{EmoryNLP} \\
    \cline{2-9}
    ~ & ES & woES & ES & woES & ES & woES & ES & woES \\
    \hline
    DialogRNN & 47.50 & 69.20 & - & - & - & - & - & -\\
    DialogXL & 55.00 & - & - & - & - & - & - & - \\
    DAG-ERC & 57.98 & 74.25 & 59.02 & 69.45 & 57.26 & 59.25 & 37.29 & 42.10 \\
    BERT-ERC (teacher) & \textbf{59.38} & \textbf{77.64} & \textbf{60.02} & \textbf{73.52} & \textbf{66.27} & \textbf{60.13} & \textbf{37.59} & \textbf{44.32} \\
    \hline
    \end{tabular}
    \caption{Test accuracy of several models with and without emotion shift (\%). (ES: emotion shift; woES: without emotion shift)}
    \label{tab:emotion shift}
\end{table*}

\subsection{Error Study}
Reviewing the prediction results, our model performs poorly in two situations, namely ambiguous emotion and emotion shift. 

For the former, BERT-ERC (teacher) fails to accurately differentiate similar emotions, such as \textit{happiness vs excited}, \textit{peaceful vs neutral}, and \textit{anger vs frustrated}. Moreover, distinguishing other emotions from \textit{neutral} in the datasets where the neutral emotion predominates is also a struggle. We believe that the proposed common-sense-based knowledge alleviates these issues to some extent, and the improved student performance also supports our idea. Moreover, future work on this issue can also attempt to incorporate psychological expertise.

\begin{figure*}[t]
  \centering
  \subfigure[]{
    \includegraphics[width=0.4\textwidth]{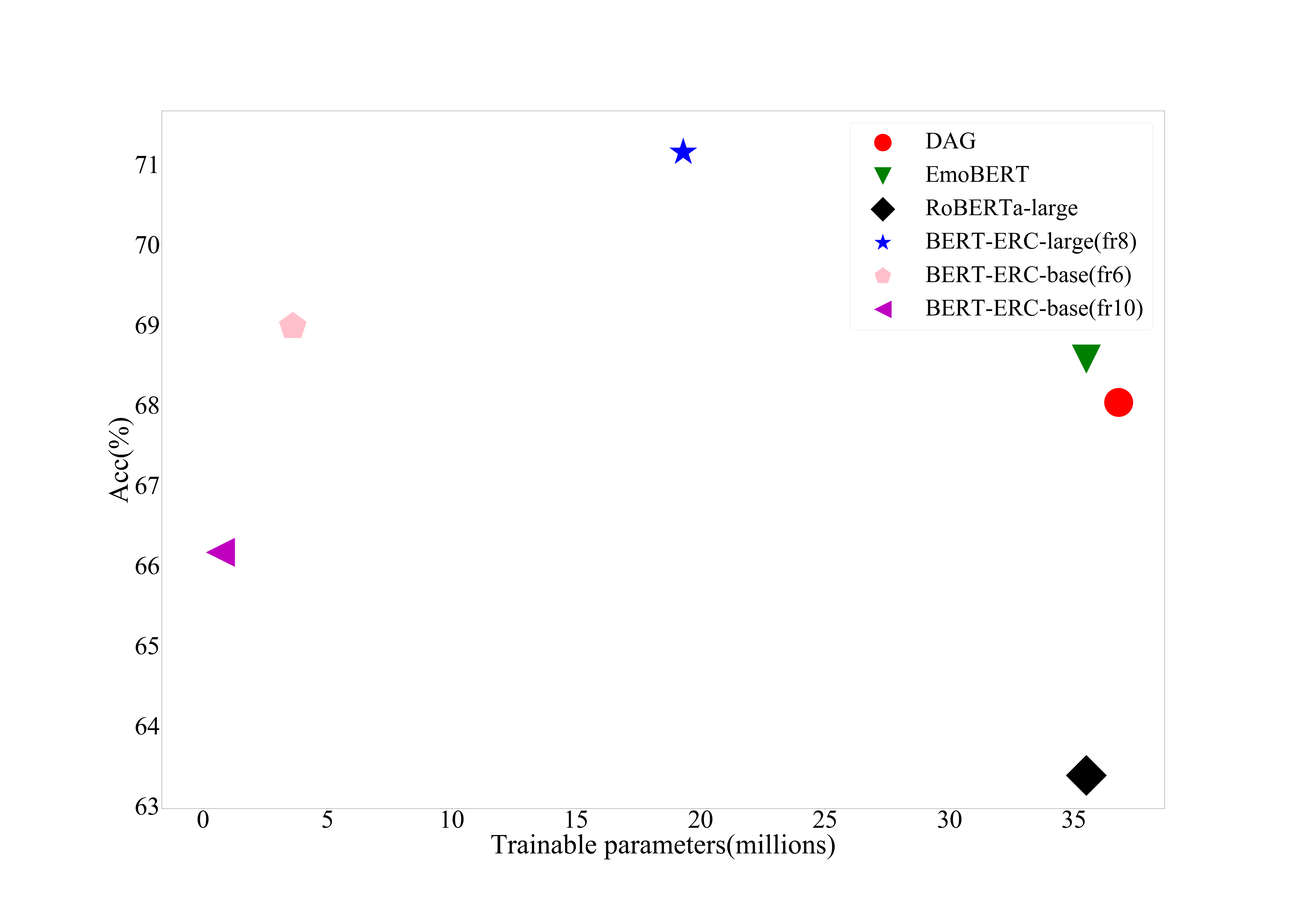}
  }
  \subfigure[]{
    \includegraphics[width=0.4\textwidth]{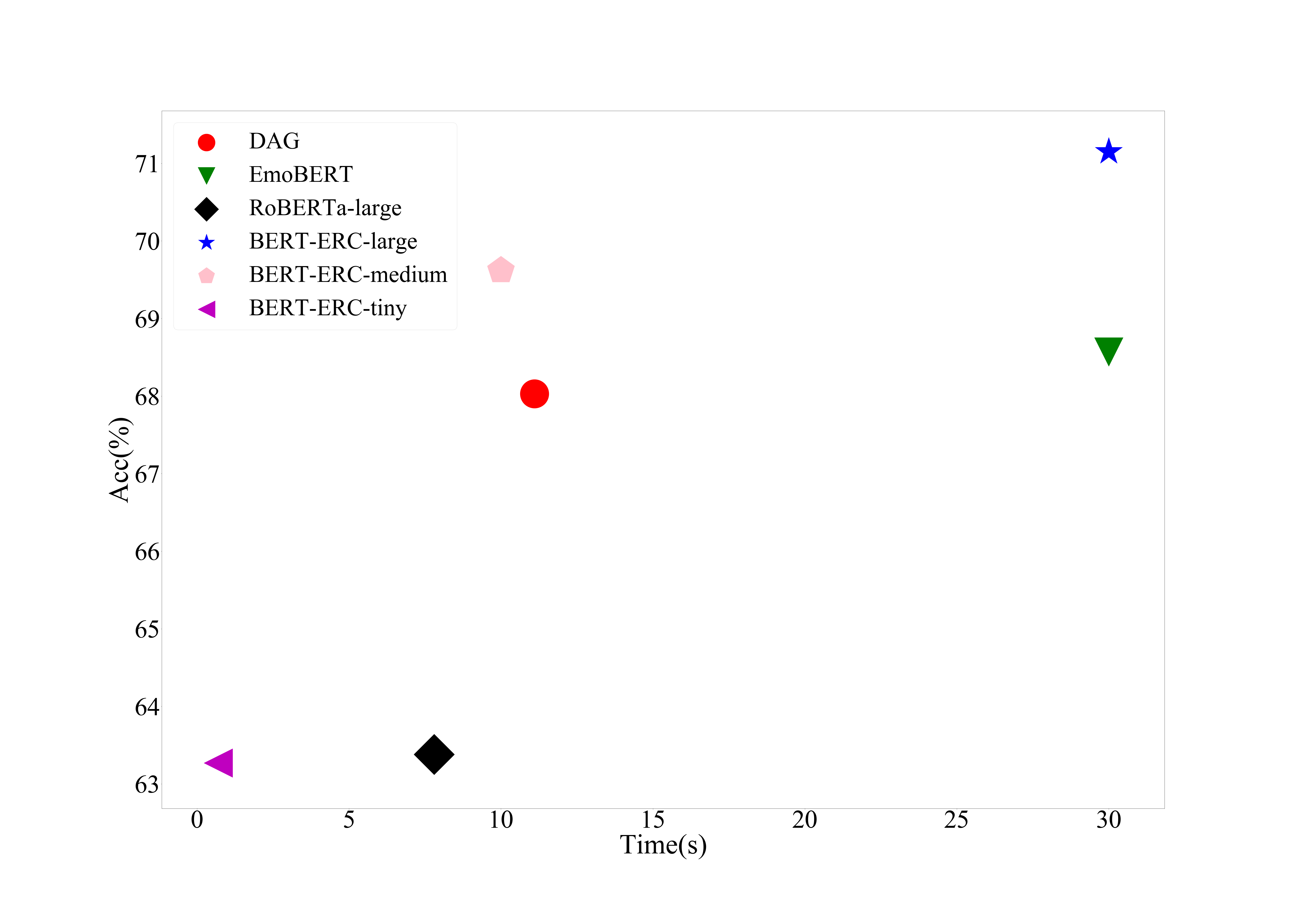}
  }
  \caption{(a) Limited resources scenario. (b) OERC scenario.}
  \label{result}
\end{figure*}

\begin{table*}[t]
    \centering
    \begin{tabular}{c|c|c|c|c|c|c}
    \hline
    \multirow{2}*{} & \multicolumn{2}{c}{Overall (\%)} \vline & \multicolumn{2}{c}{Emotion-shift (\%)} \vline & \multicolumn{2}{c}{Emotion-constant (\%)} \\
    \cline{2-7}
    ~ & MSA & SSA & MSA & SSA & MSA & SSA \\
    \hline
    IEMOCAP & 63.22	& \textbf{66.16} & \textbf{53.30} & 50.35 & 69.26 & \textbf{75.35} \\
    \hline
    MELD & 63.43 & \textbf{64.11} & 58.13 & \textbf{58.52} & 74.56 & \textbf{75.38} \\
    \hline
    DailyDialog & 54.61 & \textbf{55.70} & 51.19 & \textbf{52.84} & 49.12 & \textbf{55.73} \\
    \hline
    EmoryNLP & 35.19 & \textbf{35.72} & 33.43 & \textbf{34.03} & \textbf{45.15} & 44.32 \\
    \hline
    \end{tabular}
    \caption{Performance comparison between MSA and SSA on four datasets. (PLM: RoBERTa-base with the 10 layers frozen)}
    \label{tab:SSA and MSA}
\end{table*}

For the latter, \textit{emotion shift} is a great challenge in the ERC task, which means the emotions of two consecutive utterances from the same speaker are different. Existing ERC algorithms generally perform poorly in \textit{emotion shift}, and our model is no exception. As shown in Table~\ref{tab:emotion shift}, BERT-ERC achieves higher accuracy in recognizing samples without emotion shift than with it in most cases. Nonetheless, we still make significant progress in this scenario compared to previous works, and the comparison results are shown in Table~\ref{tab:emotion shift}. Through analysis, we believe that improving the representation ability of the model serves as the optimal solution to this problem. As the proposed paradigm builds the entire training process on the PLM, we may exploit powerful PLMs to solve this problem in the future.

\section{Comparison Experiments}

To better illustrate the advantages of the proposed paradigm, we additionally conduct comparison experiments in the three scenarios. Specifically, regarding the offline prediction, we visualize three utterances and the corresponding prediction results of BERT-ERC (teacher) and DAG-ERC on IEMOCAP dataset. For the limited resource scenario and the OERC scenario, we compare the number of trainable parameters and inference speed of several models, respectively.

\subsection{Visualization}
To intuitively show the advantages of the proposed model and the flaws of existing methods, we conduct visualization in Figure~\ref{example}. Further analysis is as follows.

Regarding utterance (a, 1), both models make the correct prediction of the speaker emotion. However, DAG-ERC fails to understand the emotion change from utterance (a, 1) to utterance (a, 2) and thus makes a wrong judgement, while our model comprehends the mood swings in the conversation via introducing the dialogue structure information and contextual information into the fine-tuning step. 

In dialogue b, Bob’s words are split into utterance (b, 0) and utterance (b, 2), which poses obstacles to analyzing the emotion state of former utterance. In fact, the “That I haven't been able to have a job.” in utterance (b, 2) implies that the emotion state of both utterances is \textit{frustrated}. However, the abstract utterance-level features in the previous paradigm make it difficult to analyze contextual information, thus leading to the wrong prediction of DAG-ERC. Differently, BERT-ERC inserts the contextual information into the PLMs and makes the correct prediction for utterance (b, 0). Moreover, in utterance (b, 3), DAG-ERC fails to distinguish the emotion state of Alice and Bob while our model addresses this issue via introducing the dialogue structure information into the PLMs.

Concerning utterance (c, 1), which contains positive “yeah, yeah, yeah” and negative “I'll have a little debt in the end.”, DAG-ERC classifies it as \textit{neutral}. As a matter of fact, Alice points out the excitement of Bob in utterance (c, 2). Unfortunately, methods following the previous paradigm cannot extract such hints, as they do not consider contextual information and dialogue structure information until the second stage. Differently, the proposed paradigm explores these two kinds of information when fine-tuning the PLMs and thus facilitates the comprehension of the interaction between the speakers.

\subsection{Trainable Parameters}
We compare the number of trainable parameters of several ERC models in Figure~\ref{result}(a). Methods following the previous paradigm, such as DAG-ERC, only utilize query utterance information when fine-tuning the PLMs, thus leading to the difficulty in analyzing contextual information and dialogue structure information. Moreover, EmoBERTa confronts the similar problem as it does not adapt the fine-tuning process to the ERC task. Thus, these algorithms make all parameters (including the parameters in the word embedding layer) trainable in the training process for better modelling capability. Differently, the proposed BERT-ERC-large (fr 8) freezes the first 8 encoder layers and the word embedding layer. As shown in Figure~\ref{result}(a), it significantly outperforms DAG-ERC and EmoBERTa with only half trainable parameters, which can be credited to the integration of query utterance information, contextual information, and dialogue structure information when fine-tuning the PLM. To meet practical needs, we use limited-scale PLMs in the limited resource scenario. As illustrated in Figure~\ref{result}(a), compared to DAG-ERC and EmoBERTa, BERT-ERC-base (fr 6) significantly reduces the trainable parameters, whereas still achieves better performance. Moreover, BERT-ERC-base (fr 10) also obtains comparable performance to these two methods with only a negligible number of trainable parameters. Overall, the proposed paradigm can be well adapted to the limited resources scenario and outperforms the methods that require a large amount of computational resources.

\subsection{Inference speed}
To evaluate the inference speed of the proposed model in the OERC scenario, we compare the prediction time of different ERC models on the test set of IEMOCAP (1623 utterances). As shown in Figure~\ref{result}(b), EmoBERTa and BERT-ERC-large take lots of time for inference, which can be ascribed to the introduced contexts when fine-tuning the PLMs. To address this issue, we propose the two-stage training strategy to utilize the time in the speaking stage. In such a manner, BERT-ERC-medium outperforms DAG-ERC in both accuracy and inference speed. Besides, we further reduce the PLM size to meet the speed requirement of several applications, such as automatic conversational agents. As shown in Figure~\ref{result}(b), BERT-ERC-tiny achieves the performance of 63.27\% and the cost time in inference is almost negligible compared to other methods. Overall, with the combination of a large-scale PLM and a tiny-scale PLM, we meet the real-time requirement in the OERC scenario at the cost of limited performance loss.

\section{Multi-Speaker Aggregation and Single-Speaker Aggregation}

To demonstrate the effectiveness of single-speaker aggregation in the resource-limited scenario, we conduct comparison experiments on four datasets using RoBERTa-base with the first 10 layers frozen as the PLM. As the results in Table~\ref{tab:SSA and MSA} show, single-speaker aggregation outperforms multi-speaker aggregation in overall weighted fscore on all datasets. Moreover, it achieves better performance in most emotion-shift cases and emotion-constant cases. Accordingly, single-speaker aggregation not only provides a boost when conversational emotions remain stable, but also focuses the model on the mood swings of the speaker, thereby improving ERC performance in the emotion-shift cases. Compared to existing methods, which use RoBERTa-large as the PLM, our approach (BERT-ERC + RoBERTa-base(fr 10) + SSA) achieves comparable performance with only 2\% trainable parameters.